\documentclass[conference]{IEEEtran} 

\usepackage{caption}
\usepackage{cite}
\usepackage{amsmath,amssymb,amsfonts}
\usepackage{algorithmic}
\usepackage{graphicx}
\usepackage{textcomp}
\usepackage{tabularx}
\usepackage{makecell}
\usepackage{multirow}
\usepackage{algorithm2e}
\RestyleAlgo{ruled}
\usepackage{hyperref}
\usepackage{balance}
\usepackage{subcaption} 

\usepackage{soul}
\usepackage{xcolor}
\usepackage{tcolorbox}

\IEEEoverridecommandlockouts    
\newcolumntype{b}{>{\hsize=1.5\hsize}X}
\newcolumntype{s}{>{\hsize=.5\hsize}X}

\title{\LARGE \bf
MTF-Grasp: A Multi-tier Federated Learning Approach\\ for Robotic Grasping
}

\author{Obaidullah Zaland, Erik Elmroth, and Monowar Bhuyan
\thanks{This work was partially supported by the Wallenberg AI, Autonomous Systems and Software Program (WASP) funded by the Knut and Alice Wallenberg Foundation via the WASP NEST project “Intelligent Cloud Robotics for Real-Time Manipulation at Scale.” The computations and data handling essential to our research were enabled by the supercomputing resource Berzelius provided by the National Supercomputer Centre at Linköping University and the gracious support of the Knut and Alice Wallenberg Foundation.}
\thanks{All the authors are with the Department of Computing Science,
        Ume\r{a} Unviersity, SE-90781 Ume\r{a}, Sweden
        {\tt\small \{ozaland,elmroth,monowar\}@cs.umu.se}}%
}

\begin{document}

\maketitle

\begin{abstract}


Federated Learning (FL) is a promising machine learning paradigm that enables participating devices to train privacy-preserved and collaborative models. FL has proven its benefits for robotic manipulation tasks. However, grasping tasks lack exploration in such settings where robots train a global model without moving data and ensuring data privacy. The main challenge is that each robot learns from data that is nonindependent and identically distributed (non-IID) and of low quantity.  This exhibits performance degradation, particularly in robotic grasping. Thus, in this work, we propose MTF-Grasp, a multi-tier FL approach for robotic grasping, acknowledging the unique challenges posed by the non-IID data distribution across robots, including quantitative skewness. MTF-Grasp harnesses data quality and quantity across robots to select a set of ``top-level'' robots with better data distribution and higher sample count. It then utilizes top-level robots to train initial seed models and distribute them to the remaining ``low-level'' robots, reducing the risk of model performance degradation in low-level robots. Our approach outperforms the conventional FL setup by up to 8\% on the quantity-skewed Cornell and Jacquard grasping datasets.



\end{abstract}



\section{INTRODUCTION}
Robotic grasping focuses on manipulators grasping objects effectively and has found potential applications in mainstream robotic tasks~\cite{zalandflcrsurvey}, including warehouse automation, search and rescue, as well as manufacturing and assembly~\cite{wang2025denoising}. Increasingly, robotic tasks necessitate a group of robots to perform similar tasks across various aforementioned applications. Thus, the need for collaboration in robotic \textit{fleets} is evident~\cite{mandi2024roco}. However, most current robotic grasping algorithms focus on stand-alone and cloud-based centralized training scenarios. In the former scenario, the robots individually train models with their local datasets, which can be used locally for inference~\cite{kleeberger2020survey}. While this scenario removes the need for communication and maintains privacy, it does not benefit collaboration and knowledge sharing among robots. Conversely, cloud-based training involves robots transferring raw data to the cloud, where a centralized model is trained. Inference in cloud-based grasping can be carried out by either the cloud or the models can be transferred to robots for local inference~\cite{mahler2016privacy}. While providing collaboration, cloud-based training risks privacy~\cite{li2024privacy} and increases communication load, as it requires data transfer to the cloud. Federated learning (FL)~\cite{mcmahan2016federated} emerges as a promising middle ground, enabling collaboration while preserving data privacy.

\begin{figure}
    \centering
    \includegraphics[width=0.8\linewidth]{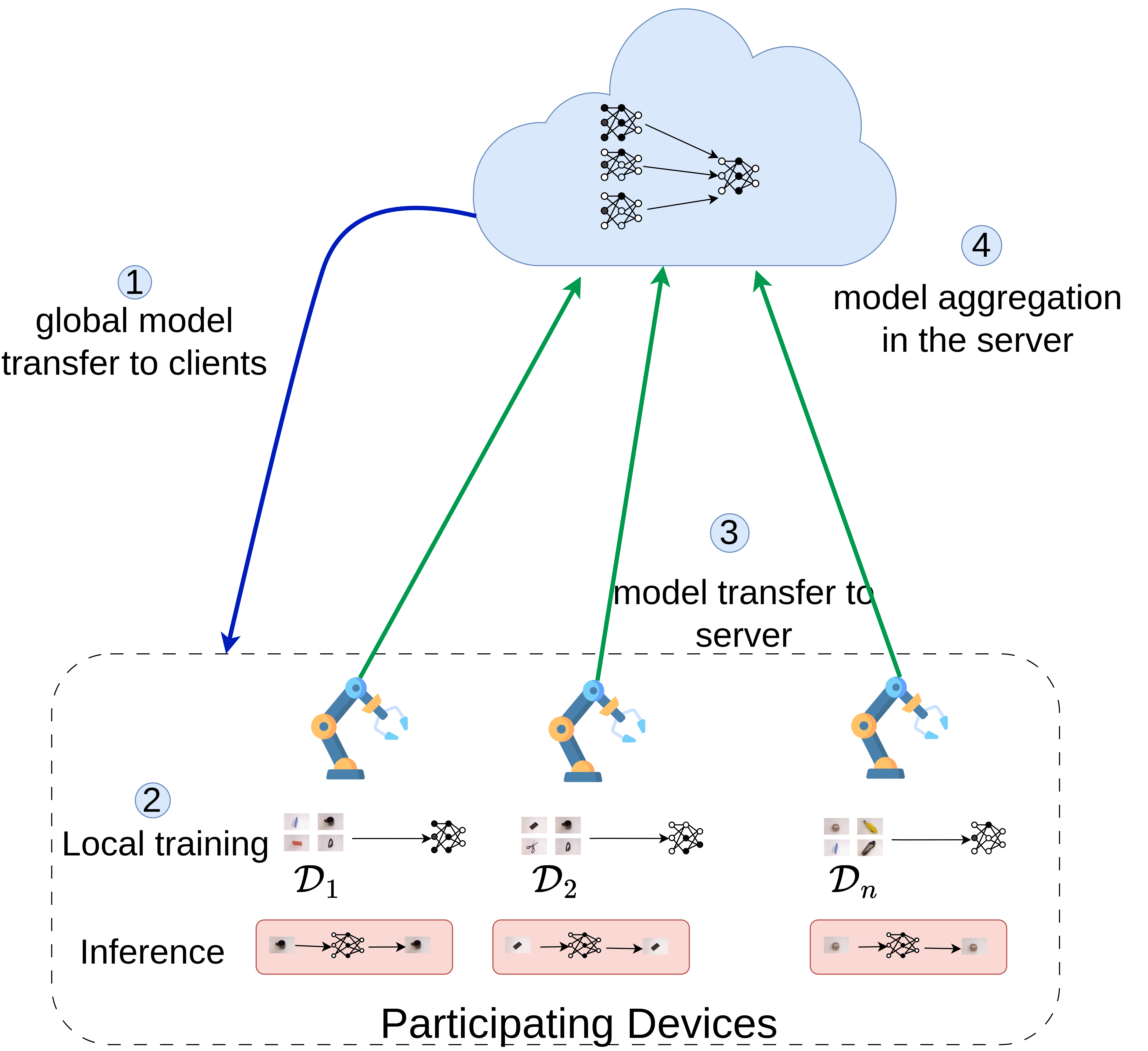}
    \caption{An overview of the federated learning pipeline.}
    \label{fig:flgrasping}
\end{figure}

FL represents a decentralized training methodology for machine learning (ML) applications, enabling participating devices to train a unified global model while keeping their data local and private~\cite{zaland2025mitigating}. Figure {\color{red}\ref{fig:flgrasping}} illustrates the FL process, where the robots train local models using their own data, which are subsequently aggregated at a \textit{server} to form the global model. The local data storage helps FL tackle privacy challenges while enabling collaborative learning. Unlike centralized ML systems, which require data transfer to a central location for model training to facilitate knowledge sharing, FL allows devices to keep their data locally while contributing to global model training, making FL ideal for data-sensitive applications. As the exposure of personal data is potent in robotic systems, FL has been increasingly applied to the domain lately~\cite{li2020graph}.

While FL enables collaboration between devices, several challenges persist, including data and system heterogeneity, communication cost, and privacy~\cite{zaland2025one}. Data heterogeneity is one of the main challenges in the robotic domain~\cite{liu2024fedcgsu}, as different robots may own \textbf{i)} data from different distributions or \textbf{ii)} different quantities of data, both cases representing non-independent and identically distributed (non-IID) data~\cite{kairouz2021advances}. Quantity-skewed data, where the amount of data across different devices is imbalanced, can lead to inaccuracy and limited generalization in FL~\cite{verma2019approaches}.
Furthermore, data quantity skewness introduces clients with limited data. Limited data can lead to local model overfitting, which is crucial in robotic manipulation tasks, as collecting new data is costly and data augmentation is challenging due to the complex nature of these tasks. To mitigate problems arising from non-IID and, specifically, quantity data skewness in robotic grasping, we propose \textbf{MTF-Grasp}. This multi-tier FL approach serves as a knowledge transfer mechanism between robots, reducing model overfitting in robots with limited data. MTF-Grasp ranks robots based on data quality and quantity, dividing them into top-level and low-level robots. It leverages the improved data distribution in top-level robots to enhance models in low-level robots. The summary of our contributions is as follows:
\begin{itemize}
\item We introduce MTF-Grasp, a novel multi-tier federated learning (FL) framework designed to address the challenges of non-IID data distributions and data scarcity commonly encountered in robotic applications.
\item We propose a client ranking mechanism within the multi-tier architecture to strategically select top-tier clients for initial model training, thereby mitigating the impact of lower-quality models from data-scarce clients.
\item We empirically evaluate the performance of MTF-Grasp against the standard (vanilla) FL approach using two distinct datasets, demonstrating its superior effectiveness.
\item We conduct a detailed communication cost analysis, showing that despite its multi-tier structure, MTF-Grasp does not introduce additional overhead compared to traditional FL setups.
\end{itemize}

\section{RELATED WORK}
This work is closely related to FL methods tailored for non-IID data and FL techniques applied to robotic manipulation. Accordingly, this section provides an overview of the current literature on the aforementioned topics.

\subsection{Federated Learning with non-IID Data}
FL presents an approach where robots collaborate on a global model while preserving the privacy of their data by keeping it on-device. The participating devices may hold data from different distributions (i.e., non-IID data); thereby, FL requires tailored methods to mitigate data heterogeneity. Numerous studies have investigated alleviating the challenges arising from non-IID data in FL settings. FedProx~\cite{li2020federated} introduces an additional regularization term to the local objective function, limiting the deviation between the regional and global models, which ensures that the averaged model is not far from the global optimum. FedNova~\cite{wang2020tackling} normalizes the local updates based on several local epochs in each device in the aggregation stage to account for the heterogeneity in the computational power of participating devices. SCAFFOLD~\cite{karimireddy2020scaffold} introduces control variates for the devices and the server, which are then used to estimate the update direction for the devices and the server.

Quantity-skewed data as a subset of non-IID data has also been investigated either independently of other non-IID cases or in association with them. In~\cite{li2022federated}, the authors have studied the effect of data quantity skew in real and synthetic datasets. Qu et al.,~\cite{qu2021experimental} investigate the impact of data skew in medical images. FedCCEA~\cite{shyn2021fedccea} presents a contribution evaluation approach from both a quality and quantity perspective. Ran et al.,~\cite{ran2021dynamic} propose dynamic reweighting of logit outputs for data imbalance cases. These studies have used model scaling, knowledge sharing, and self-balancing~\cite{duan2020self} to tackle data skewness in FL systems. 

\subsection{Federated Learning for Robotic Manipulation}
Robots are usually deployed in fleets to carry out \textit{similar} tasks; hence, FL has found potential application as a knowledge-sharing paradigm in robotic applications. SDRL~\cite{zhu2022swarm} applies decentralized FL to swarm robotics, where robots train their local actor-critic models using deep reinforcement learning and share them to create a global model. FLDDPG~\cite{na2023federated} applies FL for collective navigation in swarm robotics and reports improved communication efficiency compared to a centralized setup. Federated imitation learning~\cite{liu2020federated} proposes a novel FL-based fusion framework for cloud robotics. 

Grasping has been studied widely in centralized settings as a robotic manipulation task. Various methods, including imitation learning~\cite{pokorny2013grasp}, sampling-based learning~\cite{gou2021rgb}, and end-to-end learning~\cite{alliegro2022end} have been proposed. In FL, on the contrary, the literature on grasping is scarce. Huang et al.,~\cite{huang2023fed}, propose Fed-HANet, which proposes federated grasping for human-robot handovers. While Fed-HANet focuses on grasping, it generally compares the performance of centralized and federated grasping, overlooking data heterogeneity in FL settings. FOGL~\cite{kang2023fogl} studies grasping in federated settings, where the authors propose a combination of centralized FL settings and clustered FL settings to learn a global model. The dual process in the learning thus introduces additional time and resource complexity on the participating devices. FOGL, when considering non-IID data scenarios, does not account for the specific case of data skewness.

\begin{figure*}[!htb]
    \includegraphics[width=0.75\textwidth]{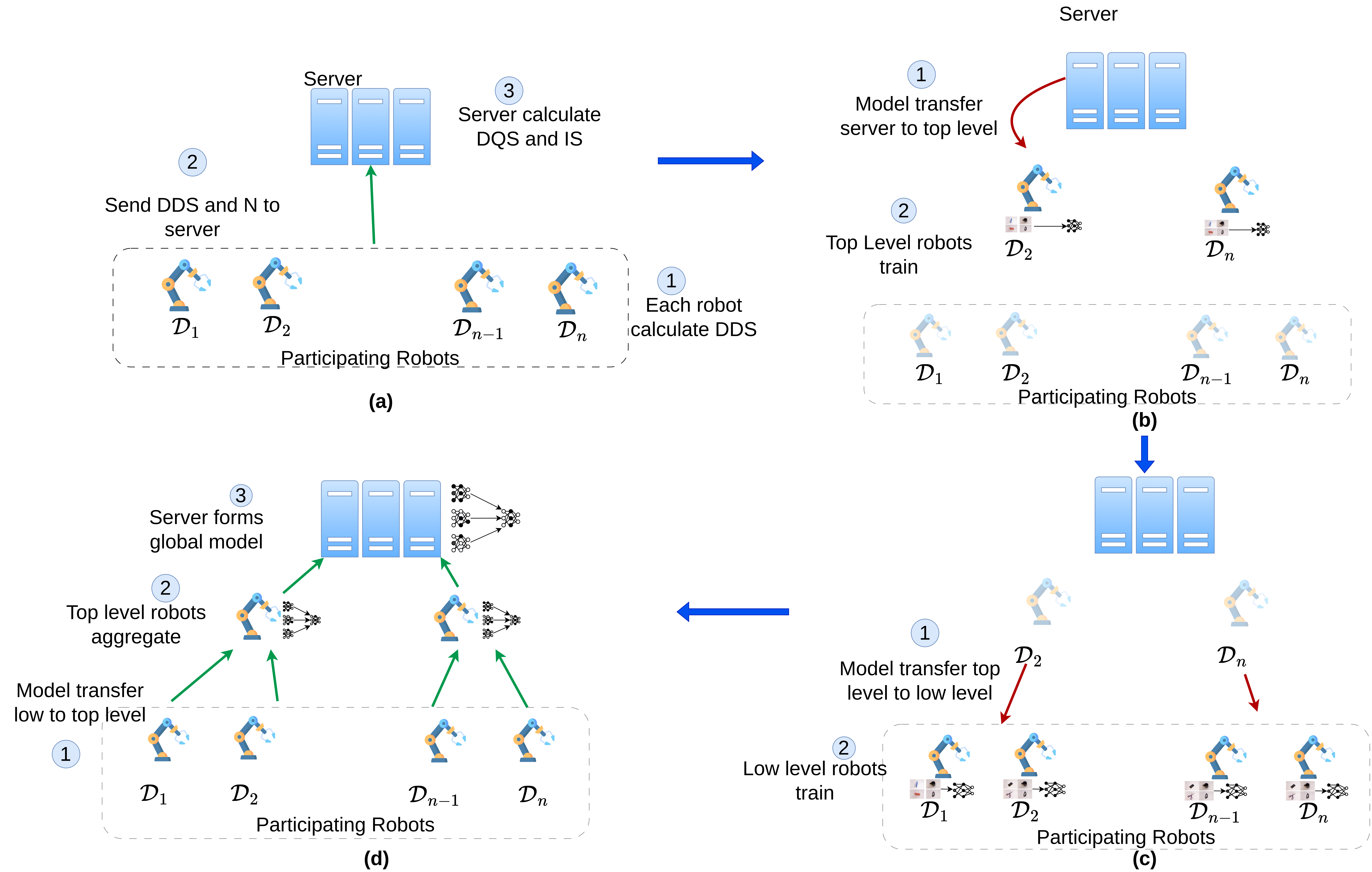}
    \centering
    \caption{MTF-Grasp process. In (a), the robots send the calculated $DDS$ score, and the server selects top-level robots after calculating $IS$ for each robot. In (b), the server sends an initialized model to be trained in top-level robots. (c) The top-level robots then send the trained models to be trained in low-level robots. (d) A two-step aggregation is carried out.}
    \label{fig:MTF-Grasp}
\end{figure*}

\section{PRELIMINARIES}
\subsection{Federated Learning}
 A centralized FL system contains a server responsible for the model aggregation and distribution alongside $n$ participating devices $\mathcal{R}$, where each device owns a local dataset $\mathcal{D}^{r\in\mathcal{R}}$, and the cumulative dataset can be represented as $\mathcal{D} = \sum_{r\in\mathcal{R}}{\mathcal{\mathcal{D}}^{r}}$.
 Generally, every training iteration in the FL process contains three different stages:

\textit{Step 1: Model Initialization}: The server initializes a global model $\mathbf{w_0}$ and distributes the initialized model to the devices.

\textit{Step 2: Local Training}: The devices train the model on their local data. In the $i$th training round, based on the global model $\mathbf{w}_i$, each client trains its local model $\mathbf{w}_i^r$ using a variation of gradient descent, as follows, and communicates the local model to the server. 

\begin{equation}\label{eq:LocalUpdate}
    \mathbf{w}_{i+1}^r = \mathbf{w}_{i}^r - \eta \nabla l(\mathbf{w}_i^r, \mathcal{D}^r)
\end{equation}
where $\eta$ denotes the learning rate, and $l(\mathbf{w}_i^r, \mathcal{D}_r)$ represents the loss for the local dataset $\mathcal{D}^r$ with weights $\mathbf{w}_i^r$.

\textit{Step 3: Model Aggregation}: After receiving the local models from the clients, the server aggregates the local models to form a global model. The simplest aggregation function is FedAvg, which averages local models as follows:

\begin{equation}\label{eq:FedAvg}
    \textbf{w}_{i+1} = \sum_{r\in\mathcal{R}}{\frac{|\mathcal{D}^r|}{{|\mathcal{D}|}} \mathbf{w}_{i+1}^r}
\end{equation}

Steps 2 and 3 are repeated until the model converges.



\begin{algorithm}[!htb]
    \caption{MTF-Grasp} \label{alg:three}
    \textbf{Input}: Local datasets $\mathcal{D}^r$, number of robots $n$, global communication rounds $E$, number of local epochs for top-level training $e_t$, number of epochs for low-level training $e_r$, learning rate $\eta$, number of data categories $m$, \\
    \label{mtf-grasp-algorithm}
    \textbf{Output}: Final grasping model $\mathbf{w}_E$\\
    \textbf{Server Initiates Training} \\
    \textbf{Robots execute:}\\  
    \ForEach{$r \in \mathcal{R}$ }  {
        Calculate $DDS_r$\\
        Send ($DDS_r$, $|\mathcal{D}^r|$) to server
    }
    \textbf{Server executes:} \\
    \ForEach{$r \in \mathcal{R}$ }  {
        Calculate $IS_r$\\
    }
        Assign top $j$ robots to $\mathcal{T}$\\
        Assign robots to $\mathcal{L}_t$ for $t \in \mathcal{T}$\\
        Initialize model $\mathbf{w}_0$\\
        Send $\mathbf{w}_0$ to $\mathcal{T}$\\

    \For{i=0,1,2,...,$E-1$}   {
        \textbf{Robots execute:}\\
        \ForEach{$t \in \mathcal{T}$}   {
            $\tilde{\mathbf{w}}_i^t \gets \mathbf{LocalUpdate}(t, \mathbf{w}_i^t, e_t$)\\
            \ForEach{$r \in \mathcal{L}_t$} {
                $\mathbf{w}_{i+1}^r \gets \mathbf{LocalUpdate}(r, \tilde{\mathbf{w}}_i^t, e_r)$
            }
            $ \mathbf{w}_{i+1}^t = \sum_{r\in \mathcal{L}_t}{\frac{|\mathcal{D}^r|}{\sum_{r\in \mathcal{L}_t} {|\mathcal{D}^r|}} \mathbf{w}_{i+1}^r}$\;
        }
        \textbf{Server Executes:}\\
        $\mathbf{w}_{i+1} = \sum_{t\in \mathcal{T}} {\frac{\sum_{r\in \mathcal{L}_t} {|\mathcal{D}^r|}}{|\mathcal{D}|}} \mathbf{w}_{i+1}^t$
    }
\end{algorithm}

\begin{algorithm}
    \caption{LocalUpdate}
    \textbf{Input}: Initial model $\mathbf{w}$ local dataset $\mathcal{D}^r$ for robot $r$, number of local rounds $e$, learning rate $\eta$, number of data categories $m$, \\
    \label{localupdate}
    \For{\text{local epoch } $k=1,2, ..., e$} {
        \For{\text{batch } $\mathbf{b} = \{x,y\} \in \mathcal{D}^r$} {
            
            $\mathbf{w}^t \gets \mathbf{w}$\\
            $\mathbf{w}^t \gets \mathbf{w}^t - \eta \nabla l(\mathbf{w}^t, \mathbf{b})$\\
            return $\mathbf{w}^t$
        }
    }
\end{algorithm}

\section{SYSTEM DESIGN}
Consider a set $\mathcal{R}$ of $n$ robots, $\mathcal{R}$, where each robot collects and learns from its local dataset $\mathcal{D}^{r\in\mathcal{R}}$, and the cumulative dataset can be represented as $\mathcal{D} = \sum_{r\in\mathcal{R}}{\mathcal{\mathcal{D}}^{r}}$. The total set of classes or categories in the dataset $\mathcal{D}$ is $\mathcal{C}$ of size $m$, and $\mathcal{D}^r_c$ represents the data of class $c\in\mathcal{C}$ belonging to robot $r\in\mathcal{R}$. We assume that the data distribution is skewed among the robots. To address the non-IID nature of the data, specifically data quantity skewness, MTF-Grasp selects a subset of robots for initial training based on the distribution of data quality and quantity. The client selection and training processes are discussed below and depicted in Figure {\color{red}\ref{fig:MTF-Grasp}}.

\subsection{Top-Level Robots Selection}
To train seed models for robots with highly non-IID data or limited data, MTF-Grasp selects $j$ robots based on their data quality and quantity as top-level robots. The criterion for calculating both is discussed below. 

\subsubsection{Data Quality} From a data quality perspective, the robots with the lowest deviation amongst the samples of different global classes are considered higher-quality robots. Consider two robots who own (3,100,2), and (52,76,88) samples for three different global classes respectively. As the deviation among the number of samples for the second robot is lower, the data quality in that robot is deemed to be higher. The data distribution score that quantifies quality is calculated as Eqn. \ref{eq:DDS} for each robot. 

\begin{equation}\label{eq:HCS}
    HCS_r = \underset{c}{\arg\max |D_c^r|} \text{ where } c\in\mathcal{C}
\end{equation}
\begin{equation}\label{eq:DDS}
    DDS_r = {\sum_{c\in\mathcal{C}} \frac{|D_c^r|}{HCS_r}} 
\end{equation}

\noindent where $\mathcal{C}$ denotes the set of classes, $|D_c^r|$ denotes the number of samples in robot $r$ of class $c$, and $HCS_r$ denotes the number of samples related to the class with the highest sample count in robot $r$.  

\subsubsection{Data Quantity} Data quantity score for each robot is quantified by \textit{how much data the robot owns with respect to other robots}, and is calculated as follows
\begin{equation}\label{eq:HRS}
    HRS = \underset{r}{\arg\max |D^r|} \text{ where } r\in\mathcal{R}
\end{equation}
\begin{equation}\label{eq:DQS}
    DQS_r = \frac{|D^r|}{HRS}
\end{equation}
\noindent where $|D^r|$ denotes the number of samples in robot $r$, and HRS denotes the number of robot samples with the highest sample count.

Both the data quality score and the data quantity score are aggregated to set an importance score (IS) for each robot, as follows.

\begin{equation}\label{eq:IS}
    IS_{r} = \lambda_{DDS} DDS_{r} + \lambda_{DQS} DQS_{r}
\end{equation}

\noindent where $\lambda_{DDS}$ and  $\lambda_{DQS}$ are hyperparameters and can be used to increase or decrease the effects of data quality and data quantity in the importance score calculation. Finally, robots are ranked based on their IS, and a set of top $j$ robots $\mathcal{T}$ are selected as top-level robots. The rest of the robots are assigned to the low-level set of a top-level robot $t$, denoted as $\mathcal{L}_t$. Note that for every top-level robot $t\in\mathcal{T}$, $t\in\mathcal{L}_t$ or every top-level robot is also included in the low-level robot subset attached to it. The initial model $\mathbf{w_0}$, initialized by the server, is sent to all robots in $\mathcal{T}$.

\subsection{MTF-Grasp - Training Process}

As the set of top-level robots $\mathcal{T}$ receive the model $\mathbf{w}_i$ in iteration $i$ from the server, they train the model for $e_t$ number of local epochs and generate $\tilde{\mathbf{w}}_i^t$, where $t \in \mathcal{T}$. The training follows some version of gradient descent as depicted in (Eqn. \ref{eq:LocalUpdate}). The intermediary model for the iteration is then sent to all low-level robots attached to the top-level robot $t$, i.e., $\mathcal{L}_t$. The training at top-level robots provides a starting point for the low-level robots, who own a smaller sample size and hence risk overfitting.

After training the intermediary model for training epoch $i$, the top-level robot sends it to all the low-level robots attached to it. Each low-level robot trains the model received from the top-level robot for $e_r$ number of epochs based on (Eqn. \ref{eq:LocalUpdate}) to generate a new model $\mathbf{w}_{i+1}^r$, where $r \in \mathcal{R}$, and send back the model to the top level robot.

The top-level robots aggregate the models received from low-level robots (in case of FedAvg, as Eqn. \ref{eq:TopFedAvg}), and send them to the server. 

\begin{equation}\label{eq:TopFedAvg}
        \textbf{w}_{i+1}^t = \sum_{r\in \mathcal{L}_t}{\frac{|\mathcal{D}^r|}{\sum_{r\in \mathcal{L}_t} {|\mathcal{D}^r|}} \mathbf{w}_{i+1}^r}
\end{equation}

The server finally aggregates all the models received from top-level robots (in case of FedAvg, as Eqn. \ref{eq:ServerFedAvg}).

\begin{equation}\label{eq:ServerFedAvg}
        \textbf{w}_{i+1} = \sum_{t\in \mathcal{T}} {\frac{\sum_{r\in \mathcal{L}_t} {|\mathcal{D}^r|}}{|\mathcal{D}|}} \mathbf{w}_{i+1}^t
\end{equation}

After the set amount of rounds, the final model is sent to all the robots, which can be used for inference. The steps of the approach are described in Algorithm \ref{mtf-grasp-algorithm}.

\begin{figure*}[ht]
    \centering
    \begin{subfigure}[b]{0.49\linewidth}
        \includegraphics[width=\linewidth]{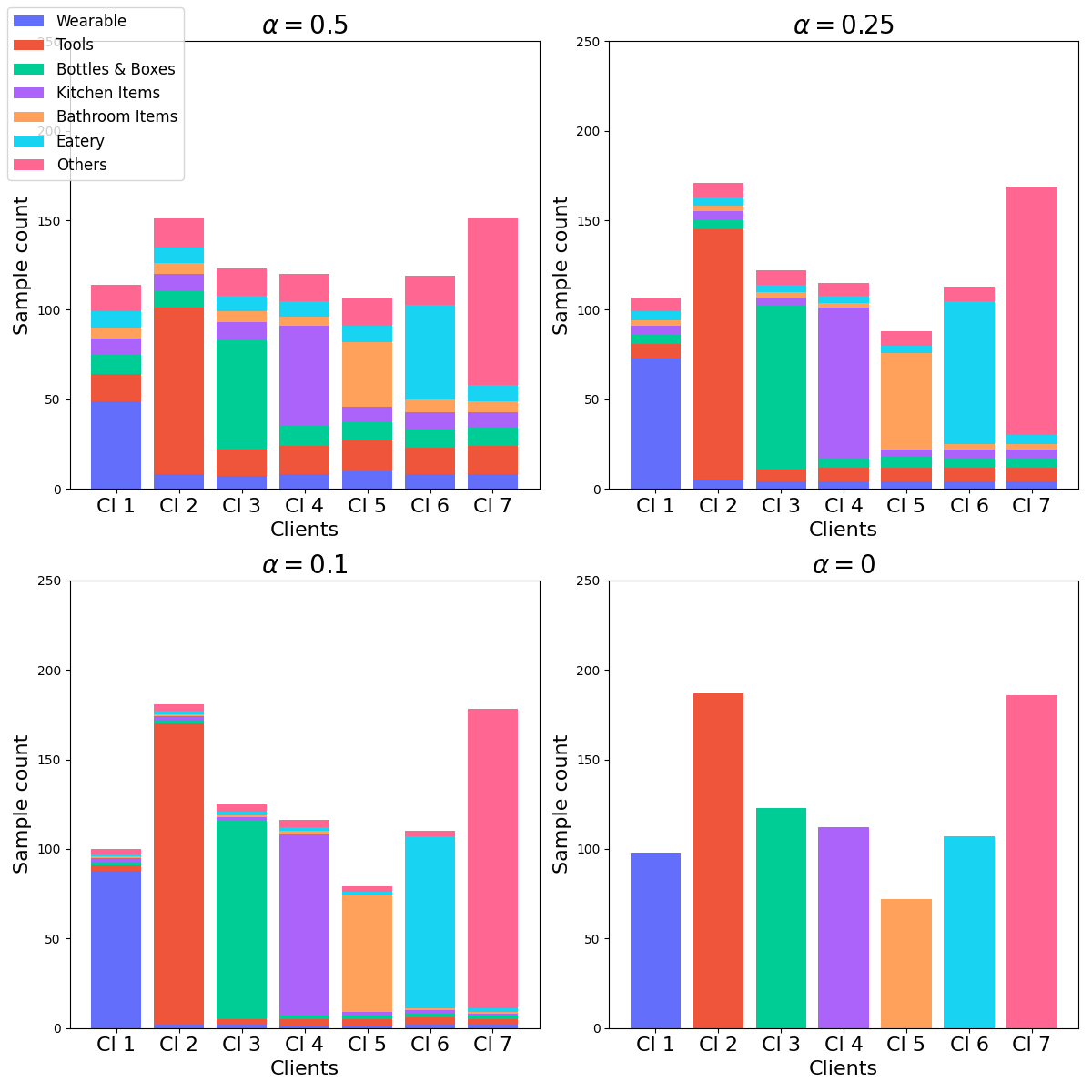}
        \caption{Class-based non-IID experiments}
        \label{fig:label_distribution}
    \end{subfigure}
    \hfill
    \begin{subfigure}[b]{0.49\linewidth}
        \includegraphics[width=\linewidth]{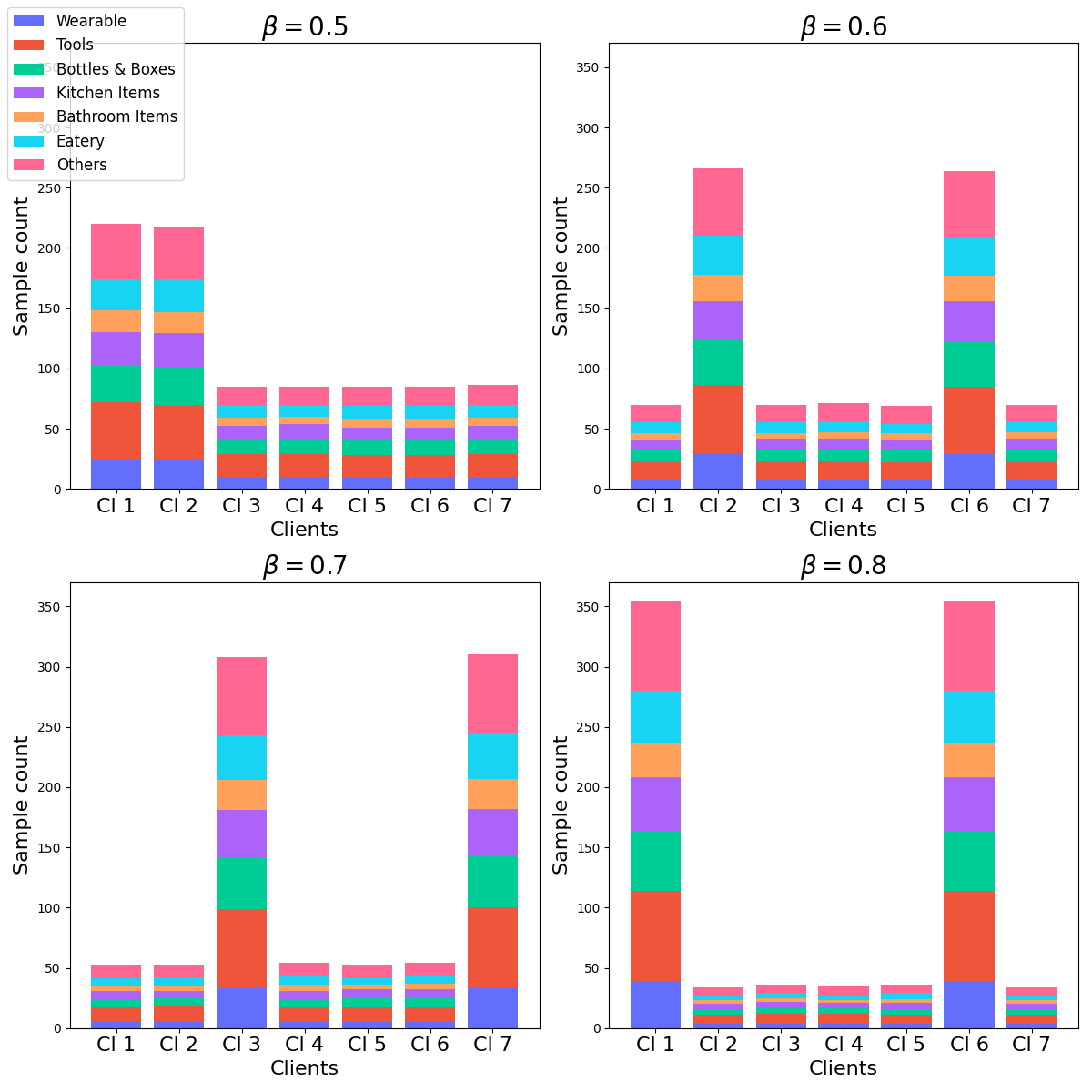}
        \caption{Quantity-skewed experiments.}
        \label{fig:skewed_distribution}
    \end{subfigure}
    \caption{Data division among clients in class-based non-IID and quantity-skewed non-IID experimental setups.}
    \label{fig:both_distributions}
\end{figure*}

\section{IMPLEMENTATION DETAILS}
\subsection{Datasets}
In this work, we use the Cornell~\cite{lenz2015deep} and Jacquard~\cite{depierre2018jacquard} grasping datasets, two of the most widely used robotic grasping datasets. Both datasets contain RGB-D images alongside annotations for the grasping areas in each image. The Cornell dataset comprises 885 RGB-D images of approximately 240 distinct real-world objects. The Jacquard dataset has around 54,000 images from around 11,500 objects. The Jacquard dataset comprises images from ShapeNet~\cite{chang2015shapenet}, a large CAD model dataset. For our experiments, we use a subset of the Jacquard datasets, which contains approximately 4000 distinct objects, hereafter denoted as Jacquard-4k.

To simulate the non-IID distribution of data, the dataset must be divided into classes, and while each object category can act as a class, this distribution would significantly dilute objects per class. Hence, we split the Cornell dataset into seven classes for the experiment's purpose. The choice of the number of classes was inspired by ~\cite{kang2023fogl}, but we added an extra class for objects that didn't fit any of the other classes. Table \ref{table:categories} contains the classes and number of samples for each class. Fig. \ref{fig:both_distributions} provides the sample distribution across clients in both class-based non-IID and quantity-skew scenarios for the Cornell dataset.

\subsection{Model Training and Hyperparameters}

This work utilizes the grasping network defined in~\cite{zhu2022sample} and the implementation provided by the authors\footnote{https://github.com/ZXP-S-works/SE2-equivariant-grasp-learning} for the main grasping network with minor adjustments and a horizontal FL setup. The experiments were conducted with $n=7$ robots, including a subset of top-level robots $j=2$. 

Accuracy is the ratio of positive grasp predictions from the test set and the number of samples in the test set. MTF-Grasp can be utilized with any aggregation function, and in this case, MTF-Grasp is coupled with FedAvg and FedNova, shown as MTF-Grasp-Avg and MTF-Grasp-Nova in the experiments. For the algorithmic hyperparameter values, we have used $e_t=5$ for a number of local iterations in top-level robots, $e_r=15$ for the number of iterations in low-level robots, while the number of global communication rounds for FL training is set to 10. Furthermore, we found that the value $0.5$ worked best for both $\lambda_{DDS}$ and $\lambda_{DQS}$.

\begin{table}[!t]
\caption{Cornell dataset division into multiple classes}
\label{table:categories}
\centering
\begin{tabularx}{0.45\textwidth}{m{1.5cm}X}
 \hline
 Dataset & Classes\\
 \hline
 Cornell(885) & wearables (98), tools(187), bottles and boxes(123), kitchen items(112), bathroom items(72), eatery(107), others(186)\\
\hline
\end{tabularx}
\end{table}

\section{RESULTS AND ANALYSIS}
We compare the effectiveness of MTF-Grasp in quantity-based non-IID and class-based non-IID scenarios individually. We also analyze MTF-Grasp's communication complexity against that of a traditional FL setup.

\begin{table*}
\caption{Accuracy of MTF-Grasp-Avg and MTF-Grasp-Nova as compared to vanilla FedAvg and FedNova settings on the test set in quantity-skewed Cornell and Jacquard datasets}
\label{tab:skewed}
\centering

\begin{tabularx}{0.95\textwidth}{m{1.9cm}m{2.75cm}*{4}{>{\centering\arraybackslash}X}}
\hline
    &  & \multicolumn{4}{c}{non-IID} \\
    \cline{3-6}
    Datasets & Algorithm & \makecell{$\beta=0.5$} & \makecell{$\beta=0.6$} & \makecell{$\beta=0.7$} & \makecell{$\beta=0.8$}  \\
    \hline
    \multirow{4}{*}{Cornell} & FedAvg & 77.14 & 77.14  & 77.03 & 74.20\\
    & FedNova & 82.85 & 84.60 & 80.00 & 80.00\\
    & MTF-Grasp-Avg & \textbf{85.72} & 84.64 & 81.45 & 81.40\\
    
    & MTF-Grasp-Nova & 85.70 & \textbf{85.70} & \textbf{83.60} & \textbf{82.80}\\
\hline
    \multirow{4}{*}{Jacquard-4k} & FedAvg & 87.16 & 86.33  & 86.50 & 83.33 \\
    & FedNova & \textbf{91.58} & 90.24  & 88.54 & 84.66\\
    & MTF-Grasp-Avg & 91.28 & 91.75  & 90.12 & 87.10\\
    & MTF-Grasp-Nova & 91.28 & \textbf{91.89}  & \textbf{90.70} & \textbf{87.40} \\
\hline
\end{tabularx}
\end{table*}

\begin{table*}
\caption{Individual robot local model performance on the test set in quantity-skewed Cornell dataset}
\label{robotbased}
\centering
\begin{tabularx}{0.95\textwidth}{m{1.75cm}m{1.5cm}m{2.75cm}*{5}{>{\centering\arraybackslash}X}}
\hline
    Datasets & & Algorithm & R3 & R4 & R5 & R6 & R7  \\
    \hline
    \multirow{4}{*}{Cornell} & \multirow{2}{*}{$\beta = 0.5$} & FedAvg & 82.80 & 78.10 & \textbf{84.28} & 76.80 & 77.10 \\
    & & MTF-Grasp-Avg & \textbf{92.80} & \textbf{84.60} & 82.80 & \textbf{84.76} & \textbf{82.80} \\
    \cline{2-8}
    & \multirow{2}{*}{$\beta = 0.8$} & FedAvg & 80.00 & \textbf{78.10} & 78.40 & 77.64 & 79.45 \\
     & & MTF-Grasp-Avg & \textbf{82.80} & 78.40 & \textbf{84.60} & \textbf{84.60} & \textbf{80.76} \\
\hline
\end{tabularx}



\end{table*}

\subsection{Quantity Skewed Division} Quantity-skewed non-IID data is when the number of samples differs significantly among participating robots. The main focus for MTF-Grasp is to prevent model performance reduction in scenarios where data quantity is skewed. To simulate different levels of data skewness in the dataset, this work uses a parameter $\beta$ to control the level of skewness. For example, when $\beta=0.7$, around $70\%$ of data from every class is divided among $j$ robots, the number of robots to be selected for top-level robots. The rest of the data is distributed among the remaining robots. As the value of $\beta$ increases and approaches $1$, the data becomes more skewed, and higher chunks of data end up in $j$ robots, leaving the other $n-j$ robots with lesser data. This experiment aims to show how MTF-Grasp compares with the vanilla FL setup as data becomes increasingly skewed.

 Table \ref{tab:skewed} shows the performance of FedAvg and FedNova coupled with MTF-Grasp compared to their contemporary vanilla implementations. MTF-Grasp consistently performs better than vanilla settings in all the scenarios when the data is skewed across robots. While the difference between FedNova and its contemporary MTF-Grasp-Nova is still significant, the difference between FedAvg and MTF-Grasp-Avg is the highest. Furthermore, MTF-Grasp, with the most straightforward aggregation function FedAvg, can perform equal to or better than the vanilla FL setup with FedNova.

Furthermore, we looked into the effect of MTF-Grasp on local robot models. In some scenarios, while the global model can perform well, individual robot models can suffer degradation. Table \ref{robotbased} shows the performance of local models from robots that were not chosen as top-level robots. For simplicity, $R1$ and $R2$ are considered to be selected as top-level robots. It can be seen that MTF-Grasp improves the quality of local robot models alongside improving global models.

\subsection{Class-based non-IID Division} Besides comparing MTF-Grasp with FedAvg in data quantity skewed scenarios, as MTF-Grasp utilizes both data quality and data quantity to select top-level robots, we carried out experiments in class-based non-IID scenarios to test MTF-Grasp. In these scenarios, the class distribution among robots is not uniform, and one robot may contain a higher number of samples from a single class. In contrast, the samples from other classes may be limited. 

To simulate class-based non-IID distribution, the data is distributed among the robots as follows: A value, in this case, $\alpha$, is used to control the level of heterogeneity in data distribution. For example, when $\alpha=0.25$, around $1-\alpha$ proportion or around $75\%$ of each class's data, is assigned to one robot, while the rest or around $25\%$ of each class's data, is distributed among the remaining six robots. As the value of $\alpha$ decreases, the level of data heterogeneity increases, and vice versa. In this case, different values of $\alpha$, including $0.5, 0.25, 0.1, \text{ and } 0$. The class distribution of the Cornell dataset, as depicted in Table \ref{table:categories}, is used to distribute the data among $n=7$ robots. Table \ref{nonIID} shows the performance of MTF-Grasp with FedAvg compared to vanilla FedAvg. While the performance of MTF-Grasp-Avg is similar to that of FedAvg with IID data, it outperforms FedAvg in non-IID data distributions. 

\begin{table}
\caption{Results comparing MTF-Grasp with FedAvg on the test set with the Cornell dataset in IID and non-IID settings}
\label{nonIID}
\begin{center}
\begin{tabularx}{0.48\textwidth}{m{2cm}*{5}{>{\centering\arraybackslash}X}}
\hline
    & & \multicolumn{4}{c}{non-IID} \\
     \cline{3-6}
     & & \multicolumn{4}{c}{$\alpha$} \\
     \cline{3-6}
     Algorithm & IID & $0.5$ &  $0.25$ & $0.1$ & $0.0$ \\
    \hline
    FedAvg & \textbf{87.14} & 82.8 & 79.3 & 77.14 & 72.13 \\
    MTF-Grasp-Avg & \textbf{87.14} & \textbf{84.8} & \textbf{83.6} & \textbf{78.2} & \textbf{76.7} \\
\hline
\end{tabularx}
\end{center}
\end{table}

\textbf{Communication Complexity:} We further compare the proposed approach's communication complexity against the traditional FL setup to show that MTF-Grasp, while a multi-tier setup, does not add any additional communication load. Consider a neural network with parameters $\theta$, and parameter size $|\theta|$. In traditional FL settings (i.e., FedAvg), the server distributes the network to all clients, resulting in a communication load of $n \times |\theta|$ parameters. As the clients send the network after training to the server, the same amount of load is incurred, resulting in a complete load of $2 \times n \times |\theta|$ in each iteration.

In MTF-Grasp, the communication between the server and $j$ top-level robots incurs a communication load of $2 \times|\theta| \times j$. The communication between top-level and low-level robots incurs a communication load of $2 \times |\theta| \times (n-j)$, as the model is only distributed to $n-j$ low-level robots. The final communication load for each training iteration is the same as traditional FL algorithms, as calculated in the following equation, where $\mathbf{C}$ denotes the communication load of MTF-Grasp.
\vspace{-0.65cm}

\begin{align*}
        \mathbf{C} & = 2 \times |\theta| \times j + 2 \times |\theta| \times (n-j) \\
            & = 2 \times n \times |\theta|
\end{align*}

\section{DISCUSSION AND LIMITATIONS}

\paragraph{Low-level robots assignment} In a robotic setup, the low-level robots might be assigned to top-level robots based on some pre-defined criteria (e.g., geographical co-location or system homogeneity).

\paragraph{Computational complexity in FL grasping} Grasping is a complex task and requires computational resources to learn models concerning it. In a centralized setup, the server usually possesses the computational capabilities that the individual robots might lack in a federated setup. Current federated grasping works, including ours, do not consider computational efficiency.

\section{CONCLUSION}
In this work, we proposed MTF-Grasp, a multi-tier federated learning approach for robotic grasping. MTF-Grasp addresses data heterogeneity, specifically data quantity skewness, where the number of samples across robots is highly skewed. MTF-Grasp utilizes the statistical data quality and quantity of the robots to select a set of top-level robots for training initial seed models of the remaining low-level robots, assigning them to a top-level robot. This enables knowledge transfer from robots with higher sample sizes and better distributions to the remaining robots, preventing model degradation. MTF-Grasp-Avg and MTF-Grasp-Nova, which utilize FedAvg and FedNova as aggregation functions, outperform vanilla FedAvg and FedNova. In quantity-skewed scenarios, MTF-Grasp outperforms the vanilla FL approach by up to 8\% in grasping accuracy. Additionally, MTF-Grasp-Avg outperforms FedAvg in class-based non-IID scenarios. 


\balance
\bibliography{IEEEbibtex}

\begin{thebibliography}{10}
\providecommand{\url}[1]{#1}
\csname url@rmstyle\endcsname
\providecommand{\newblock}{\relax}
\providecommand{\bibinfo}[2]{#2}
\providecommand\BIBentrySTDinterwordspacing{\spaceskip=0pt\relax}
\providecommand\BIBentryALTinterwordstretchfactor{4}
\providecommand\BIBentryALTinterwordspacing{\spaceskip=\fontdimen2\font plus
\BIBentryALTinterwordstretchfactor\fontdimen3\font minus \fontdimen4\font\relax}
\providecommand\BIBforeignlanguage[2]{{%
\expandafter\ifx\csname l@#1\endcsname\relax
\typeout{** WARNING: IEEEtran.bst: No hyphenation pattern has been}%
\typeout{** loaded for the language `#1'. Using the pattern for}%
\typeout{** the default language instead.}%
\else
\language=\csname l@#1\endcsname
\fi
#2}}

\bibitem{zalandflcrsurvey}
O.~Zaland, C.~Nguyen, F.~T. Pokorny, and M.~Bhuyan, ``Federated learning for large-scale cloud robotic manipulation: Opportunities and challenges,'' in \emph{2025 IEEE International Conference on Machine Learning and Cybernetics (ICMLC)}.\hskip 1em plus 0.5em minus 0.4em\relax IEEE, 2025.

\bibitem{wang2025denoising}
H.~Wang, X.~Zhong, K.~Liu, F.~Liu, and W.~Zhang, ``Denoising and adaptive online vertical federated learning for sequential multi-sensor data in industrial internet of things,'' \emph{arXiv preprint arXiv:2501.01693}, 2025.

\bibitem{mandi2024roco}
Z.~Mandi, S.~Jain, and S.~Song, ``Roco: Dialectic multi-robot collaboration with large language models,'' in \emph{2024 IEEE International Conference on Robotics and Automation (ICRA)}.\hskip 1em plus 0.5em minus 0.4em\relax IEEE, 2024, pp. 286--299.

\bibitem{kleeberger2020survey}
K.~Kleeberger, R.~Bormann, W.~Kraus, and M.~F. Huber, ``A survey on learning-based robotic grasping,'' \emph{Current Robotics Reports}, vol.~1, pp. 239--249, 2020.

\bibitem{mahler2016privacy}
J.~Mahler, B.~Hou, S.~Niyaz, F.~T. Pokorny, R.~Chandra, and K.~Goldberg, ``Privacy-preserving grasp planning in the cloud,'' in \emph{2016 IEEE International Conference on Automation Science and Engineering (CASE)}.\hskip 1em plus 0.5em minus 0.4em\relax IEEE, 2016, pp. 468--475.

\bibitem{li2024privacy}
M.~Li, W.~Ding, and D.~Zhao, ``Privacy risks in reinforcement learning for household robots,'' in \emph{2024 IEEE International Conference on Robotics and Automation (ICRA)}.\hskip 1em plus 0.5em minus 0.4em\relax IEEE, 2024, pp. 5148--5154.

\bibitem{mcmahan2016federated}
H.~B. McMahan, F.~Yu, P.~Richtarik, A.~Suresh, D.~Bacon, \emph{et~al.}, ``Federated learning: Strategies for improving communication efficiency,'' in \emph{Proceedings of the 29th Conference on Neural Information Processing Systems (NIPS), Barcelona, Spain}, 2016, pp. 5--10.

\bibitem{zaland2025mitigating}
O.~Zaland, Y.~Onur, and M.~Bhuyan, ``Mitigating data heterogeneity with multi-tier federated gan,'' in \emph{International Conference on Neural Information Processing}.\hskip 1em plus 0.5em minus 0.4em\relax Springer, 2025, pp. 225--239.

\bibitem{li2020graph}
Q.~Li, F.~Gama, A.~Ribeiro, and A.~Prorok, ``Graph neural networks for decentralized multi-robot path planning,'' in \emph{2020 IEEE/RSJ International Conference on Intelligent Robots and Systems (IROS)}.\hskip 1em plus 0.5em minus 0.4em\relax IEEE, 2020, pp. 11\,785--11\,792.

\bibitem{zaland2025one}
O.~Zaland, S.~Jin, F.~T. Pokorny, and M.~Bhuyan, ``One-shot federated learning with classifier-free diffusion models,'' 2025.

\bibitem{liu2024fedcgsu}
H.~Liu, L.~Feng, and M.~Mei, ``Fedcgsu: Client grouping based on similar uncertainty for non-iid federated learning,'' in \emph{2024 IEEE International Conference on Systems, Man, and Cybernetics (SMC)}.\hskip 1em plus 0.5em minus 0.4em\relax IEEE, 2024, pp. 1773--1778.

\bibitem{kairouz2021advances}
P.~Kairouz, H.~B. McMahan, B.~Avent, A.~Bellet, M.~Bennis, A.~N. Bhagoji, K.~Bonawitz, Z.~Charles, G.~Cormode, R.~Cummings, \emph{et~al.}, ``Advances and open problems in federated learning,'' \emph{Foundations and Trends{\textregistered} in Machine Learning}, vol.~14, no. 1--2, pp. 1--210, 2021.

\bibitem{verma2019approaches}
D.~C. Verma, G.~White, S.~Julier, S.~Pasteris, S.~Chakraborty, and G.~Cirincione, ``Approaches to address the data skew problem in federated learning,'' in \emph{Artificial Intelligence and Machine Learning for Multi-Domain Operations Applications}, vol. 11006.\hskip 1em plus 0.5em minus 0.4em\relax SPIE, 2019, pp. 542--557.

\bibitem{li2020federated}
T.~Li, A.~K. Sahu, M.~Zaheer, M.~Sanjabi, A.~Talwalkar, and V.~Smith, ``Federated optimization in heterogeneous networks,'' \emph{Proceedings of Machine learning and systems}, vol.~2, pp. 429--450, 2020.

\bibitem{wang2020tackling}
J.~Wang, Q.~Liu, H.~Liang, G.~Joshi, and H.~V. Poor, ``Tackling the objective inconsistency problem in heterogeneous federated optimization,'' \emph{Advances in neural information processing systems}, vol.~33, pp. 7611--7623, 2020.

\bibitem{karimireddy2020scaffold}
S.~P. Karimireddy, S.~Kale, M.~Mohri, S.~Reddi, S.~Stich, and A.~T. Suresh, ``Scaffold: Stochastic controlled averaging for federated learning,'' in \emph{International conference on machine learning}.\hskip 1em plus 0.5em minus 0.4em\relax PMLR, 2020, pp. 5132--5143.

\bibitem{li2022federated}
Q.~Li, Y.~Diao, Q.~Chen, and B.~He, ``Federated learning on non-iid data silos: An experimental study,'' in \emph{2022 IEEE 38th ICDE}.\hskip 1em plus 0.5em minus 0.4em\relax IEEE, 2022, pp. 965--978.

\bibitem{qu2021experimental}
L.~Qu, N.~Balachandar, and D.~L. Rubin, ``An experimental study of data heterogeneity in federated learning methods for medical imaging,'' \emph{arXiv preprint arXiv:2107.08371}, 2021.

\bibitem{shyn2021fedccea}
S.~K. Shyn, D.~Kim, and K.~Kim, ``Fedccea: A practical approach of client contribution evaluation for federated learning,'' \emph{arXiv preprint arXiv:2106.02310}, 2021.

\bibitem{ran2021dynamic}
X.~Ran, L.~Ge, and L.~Zhong, ``Dynamic margin for federated learning with imbalanced data,'' in \emph{IJCNN}.\hskip 1em plus 0.5em minus 0.4em\relax IEEE, 2021, pp. 1--8.

\bibitem{duan2020self}
M.~Duan, D.~Liu, X.~Chen, R.~Liu, Y.~Tan, and L.~Liang, ``Self-balancing federated learning with global imbalanced data in mobile systems,'' \emph{IEEE Trans. on Parallel and Distributed Systems}, vol.~32, no.~1, pp. 59--71, 2020.

\bibitem{zhu2022swarm}
X.~Zhu, F.~Zhang, and H.~Li, ``Swarm deep reinforcement learning for robotic manipulation,'' \emph{Procedia Computer Science}, vol. 198, pp. 472--479, 2022.

\bibitem{na2023federated}
S.~Na, T.~Rou{\v{c}}ek, J.~Ulrich, J.~Pikman, T.~s~Krajn{\'\i}k, B.~Lennox, and F.~Arvin, ``Federated reinforcement learning for collective navigation of robotic swarms,'' \emph{IEEE Trans on Cognitive and Developmental Systems}, 2023.

\bibitem{liu2020federated}
B.~Liu, L.~Wang, M.~Liu, and C.-Z. Xu, ``Federated imitation learning: A novel framework for cloud robotic systems with heterogeneous sensor data,'' \emph{IEEE Robotics and Automation Letters}, vol.~5, no.~2, pp. 3509--3516, 2020.

\bibitem{pokorny2013grasp}
F.~T. Pokorny, K.~Hang, and D.~Kragic, ``Grasp moduli spaces.'' in \emph{Robotics: Science and Systems}, 2013.

\bibitem{gou2021rgb}
M.~Gou, H.-S. Fang, Z.~Zhu, S.~Xu, C.~Wang, and C.~Lu, ``Rgb matters: Learning 7-dof grasp poses on monocular rgbd images,'' in \emph{ICRA}.\hskip 1em plus 0.5em minus 0.4em\relax IEEE, 2021, pp. 13\,459--13\,466.

\bibitem{alliegro2022end}
A.~Alliegro, M.~Rudorfer, F.~Frattin, A.~Leonardis, and T.~Tommasi, ``End-to-end learning to grasp via sampling from object point clouds,'' \emph{IEEE Robotics and Automation Letters}, vol.~7, no.~4, pp. 9865--9872, 2022.

\bibitem{huang2023fed}
C.-I. Huang, Y.-Y. Huang, J.-X. Liu, Y.-T. Ko, H.-C. Wang, K.-H. Chiang, and L.-F. Yu, ``Fed-hanet: Federated visual grasping learning for human robot handovers,'' \emph{IEEE Robotics and Automation Letters}, 2023.

\bibitem{kang2023fogl}
S.-K. Kang and C.~Choi, ``Fogl: Federated object grasping learning,'' in \emph{2023 IEEE International Conference on Robotics and Automation (ICRA)}.\hskip 1em plus 0.5em minus 0.4em\relax IEEE, 2023, pp. 5851--5857.

\bibitem{lenz2015deep}
I.~Lenz, H.~Lee, and A.~Saxena, ``Deep learning for detecting robotic grasps,'' \emph{The International Journal of Robotics Research}, vol.~34, no. 4-5, pp. 705--724, 2015.

\bibitem{depierre2018jacquard}
A.~Depierre, E.~Dellandr{\'e}a, and L.~Chen, ``Jacquard: A large scale dataset for robotic grasp detection,'' in \emph{2018 IEEE/RSJ International Conference on Intelligent Robots and Systems (IROS)}.\hskip 1em plus 0.5em minus 0.4em\relax IEEE, 2018, pp. 3511--3516.

\bibitem{chang2015shapenet}
A.~X. Chang, T.~Funkhouser, L.~Guibas, P.~Hanrahan, Q.~Huang, Z.~Li, S.~Savarese, M.~Savva, S.~Song, H.~Su, \emph{et~al.}, ``Shapenet: An information-rich 3d model repository,'' \emph{arXiv preprint arXiv:1512.03012}, 2015.

\bibitem{zhu2022sample}
X.~Zhu, D.~Wang, O.~Biza, G.~Su, R.~Walters, and R.~Platt, ``Sample efficient grasp learning using equivariant models,'' \emph{arXiv preprint arXiv:2202.09468}, 2022.

\end{thebibliography}

\end{document}